\documentclass[journal]{IEEEtran}
\usepackage{cite}

\ifCLASSINFOpdf
  \usepackage[pdftex]{graphicx}
  \DeclareGraphicsExtensions{.pdf,.jpeg,.png}
\else
\fi
\usepackage{amsmath}
\usepackage{amsfonts}

\usepackage{algorithmic}

\usepackage{booktabs} 

\ifCLASSOPTIONcompsoc
 \usepackage[caption=false,font=normalsize,labelfont=sf,textfont=sf]{subfig}
\else
 \usepackage[caption=false,font=footnotesize]{subfig}
\fi
\begin{document}
%
\title{Pixel-wise Regression: 3D Hand Pose Estimation via Spatial-form Representation and Differentiable Decoder
\thanks{The authors are with the State Key Laboratory of Robotics and System, Harbin Institute of Technology, Harbin 150001, China (e-mail: zfhhit@hit.edu.cn)}
}
%
%
%

\author{Xingyuan~Zhang,
        Fuhai~Zhang
        }

%
%

\markboth{Journal of IEEE Trans on Image Processing, ~Vol.~XX, No.~XX, XXXXX~2019}%
{Shell \MakeLowercase{\textit{et al.}}: Bare Demo of IEEEtran.cls for IEEE Journals}
%



\maketitle

\begin{abstract}
  3D Hand pose estimation from a single depth image is an essential topic in computer vision and human-computer interaction. 
  Although the rising of deep learning method boosts the accuracy a lot, the problem is still hard to solve due to the complex structure of the human hand. 
  Existing methods with deep learning either lose spatial information of hand structure or lack a direct supervision of joint coordinates. 
  In this paper, we propose a novel Pixel-wise Regression method, which use spatial-form representation (SFR) and differentiable decoder (DD) to solve the two problems. 
  To use our method, we build a model, in which we design a particular SFR and its correlative DD which divided the 3D joint coordinates into two parts, 
  plane coordinates and depth coordinates and use two modules named Plane Regression (PR) and Depth Regression (DR) to deal with them respectively. 
  We conduct an ablation experiment to show the method we proposed achieve better results than the former methods. 
  We also make an exploration on how different training strategies influence the learned SFRs and results. 
  The experiment on three public datasets demonstrates that our model is comparable with the existing state-of-the-art models and in one of them our model can reduce mean 3D joint error by 25\%.
\end{abstract}


%
\IEEEpeerreviewmaketitle

\section{Introduction}
\label{Introduction}
%
%
%
%

 

\begin{figure*}[!t]
  \centering
  \includegraphics[width=\linewidth]{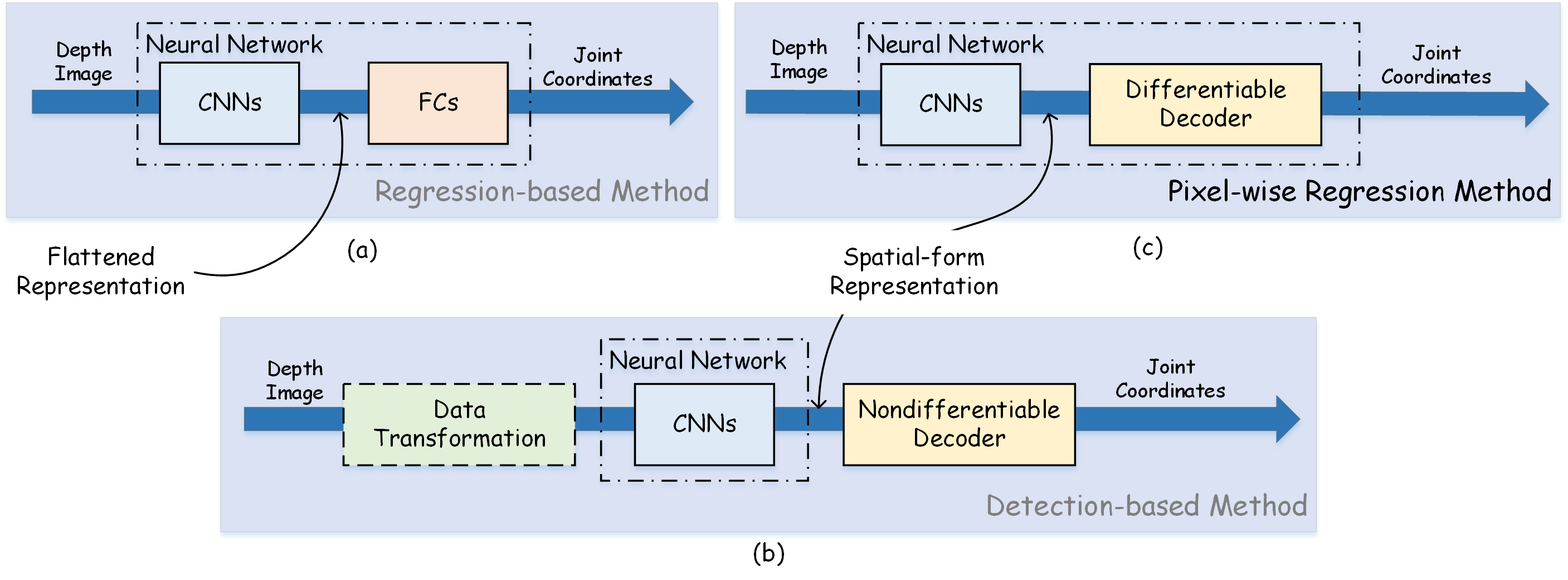}
  \caption{
    Three types of 3D hand pose estimation method. 
    (a) shows the regression-based method which uses CNNs to extract features and then uses FCs to regress the joint coordinates. 
    The flatten operation between CNNs and FCs destroy the spatial information. 
    (b) shows the detection-based method which uses CNNs to extract a designed spatial-form representation then 
    uses a nondifferentiable decoder outside the neural netwoek to convert the SFR to the joint coordinates. 
    Some models may perform a data transformation to convert the input depth image to another form like voxel map. 
    (c) shows our proposed Pixel-wise Regression method which is a combination of the former two methods. 
    We use CNNs to extract a designed spatial-form representation like the detection-based method, 
    and enables the direct supervision of 3D coordinates like regression-based method by 
    putting a differentiable decoder inside the neural network.
  }
  \label{three_types}    
\end{figure*}

Hand pose estimation is always an essential topic in computer vision and human-computer interaction\cite{vogel2015an} 
since the human hand is one of the most important things to understand the intention of human, 
and how humans interact with another object. 
This field is greatly boosted by the arising of deep learning and availability of depth cameras 
such as Microsoft Kinect and Intel Realsense\cite{iii2018depth}. 
However, the problem is still very hard to solve, 
because the complex structure of human hands, along with variety of view-point, causes severe self-occlusion in the image.

3D Hand pose estimation based on single-frame depth images mostly uses convolution neural networks (CNNs) based methods\cite{yuan2018depth}. 
The most popular method is regression-based method, which treat hand pose estimation as an end-to-end learning problem. 
That is, they build a gigantic neural network to regress the 3D coordinates of each joint point in an end-to-end fashion and let the neural network find out what features it should learn. 
The obvious defect of this regression-based method is that the output of CNNs must be flattened to go through full-connection (FC) layers to get the joint coordinates. 
This flatten operation completely destroy the spatial information in the original image, 
so it is unfavorable for recognizing such a complex structure as the hand and result in a highly nonlinear regression problem. 

In order to preserve the spatial information of human hands, the most commonly used method is the detection-based method with heatmaps, 
which also has a lot of applications in other similar fields such as human pose estimation. 
The detection-based method uses fully convolution network (FCN) to maintain the spatial information, 
and instead of joint coordinates they design an encoded spatial-form representation (SFR) as the learning target. 
Then a decoder should be used to convert the encoded SFR back to joint coordinates. 
Heatmaps (or sometimes called believe maps\cite{wei2016convolutional}) is the most commonly used and the most native choice for this encoded SFR. 
Research\cite{yuan2018depth} shows that detection-based method has a higher accuracy than the regression-based method.

However, heatmap is only an approximate representation for 2D coordinates in plane space, 
so it is not capable to deal with 3D coordinates even when the depth image is provided in 3D hand pose estimation problem. 
Because the depth image is only a projection of the real 3D data, 
there exists an offset between actual joint depth value and the value on the depth image. 
This offset can vary a lot among different joints and different gestures for the severe self-occlusion and heavy noise in depth image. 
Therefore, in order to use detection-based method in 3D hand pose estimation, 
researchers either come up with additional representations along with heatmap or 
transform the depth image back to the 3D space then use 3D heatmap or other representations. 

Although the detection-based method maintains the spatial information with SFR, it also has defects. 
First, as we talk above, the depth image is only a projection of the real 3D data, 
so it is nontrivial to define such a transformation that perfectly reconstructs the original 3D data from a single depth image. 
Second, the detection-based method lacks a direct supervision from the 3D coordinates. 
Unlike the regression-based method, in detection-based method, we supervise SFR instead of 3D coordinates. 
Since the way we build SFR is essentially encoding a low-dimensional data into high-dimension, the SFR must contain some redundancy. 
The decoder that deals with these redundancies are always non-linear and nondifferentiable in the most of time, thus has to be put outside the neural network. 
This nondifferentiable structure disable the direct supervision form the 3D coordinates and 
build a gap between what the neural network learns and what we really want. 
Like any complex non-linear system, there is no guarantee that a slight error in the SFR won’t result in a dramatic change in 3D coordinates. 
The 3D coordinates just make the situation worse because, in plane space, we need more than one form of SFR to encode the 3D coordinates. 
Since the decoder is outside the neural network, we cannot directly use the 3D coordinates to supervise the learning. 
Without such supervision, the SFRs may not corporate with each other.

In this paper, in order to overcome the defects of existing method we state above, 
we integrate existing methods and propose a Pixel-wise Regression method, 
which uses spatial-form representation to maintain the spatial information and 
enables the direct supervision of 3D coordinates by putting a differentiable decoder (DD) inside the neural network. 
To use our method, we build a particular model, 
in which we design a particular SFR and its correlative DD which divided the 3D coordinates into two parts, 
plane coordinates and depth coordinates and use two modules named Plane Regression (PR) and Depth Regression (DR) to deal with them respectively. 
The relationship among our method and regression-based method and detection-based method is shown in Fig\@. \ref{three_types}.

The contributions of this work are summarized as follows:  

\begin{itemize}
  \item We propose a novel method for 3D pose estimation called Pixel-wise regression, 
  which is a combination of existing two popular methods. 
  Our method uses SFR of 3D coordinates in plane space to maintain the spatial information, 
  and enables the direct supervision of 3D coordinates by putting a DD inside the neural network.
  \item We design a particular SFR and its correlative DD to 
  build a model for 3D hand pose estimation problem from single depth image. 
  We divided the 3D coordinates into two parts, 
  plane coordinates and depth coordinates and use two modules named PR and DR to deal with them respectively.
  \item We implement several baseline models and conduct an ablation experiment to 
  show that our method can achieve better results than former methods. 
  We also study how different training strategies influence the learned SFRs and provide meaningful insights. 
  The experiment on three public datasets shows that our model can achieve comparable result with the state-of-the-art models.
\end{itemize}

The remainder of the paper is organized as follows. 
In Section \ref{Related_Work}, we give a brief review of the regression-based and detection-based methods and 
point out the main differences between our models and their models. 
In Section \ref{METHODOLOGY}, we formalize the problem and describe our proposed Pixel-wise Regression method. 
We show how we design the SFR and its correlative DD to fulfill the requirement of our method. 
We also demonstrate other modules we use in our model and the training details. 
In Section \ref{Experiment}, we designed our ablation experiments to analyze our method and performed experiments 
on three challenging public datasets, MSRA, ICVL and HAND17. 
Experiments show that our model can achieve state-of-the-art levels on the test set. 
Finally, we make a conclusion and discuss future potential of our method in Section \ref{Conclusion}.


\section{Related work}
\label{Related_Work}

In this section, we will review some related works of our proposed model. 
Firstly, we will review the regression-based method, which is the most popular method in hand pose estimation. 
Secondly, we talk about the detection-based method that, in general, 
make a more accurate prediction than the regression-based method but lack the direct supervision from 3D coordinates. 
Our Pixel-wise regression method is a combination of these two methods.

\subsection{Regression-based Method}

Regression-based method\cite{tang2017latent, ge2019real, chen2018shpr, hu2019a, wu2018context, chen2018learning, oberweger2017deepprior, zhou2016model, chen2017pose, wu2018handmap} is widely used in hand pose estimation field. 
It treats hand pose estimation as an end-to-end problem and regresses the 3D coordinate of each joint directly. 
Some of these methods have learned the low-dimensional space or latent space representation of the hand location. 
For example, Oberweger, M et al.\cite{oberweger2017deepprior} believe that joint vectors can be regarded as some low-dimensional spatial representations. 
They use a special bottleneck structure to force the network to learn such pattern, 
but since this low-dimensional space is originally a kind of approximate expression, its effect is not good. 
X. Zhou et al.\cite{zhou2016model} used the joint angle as a hidden variable to learn angle information in the network and 
convert the angle into 3D coordinates through a predefined hand model. This model requires a pre-defined hand model, lacking generalization ability for new samples. 
Both of these models have destroyed the spatial information with a flattened representation. 
There are also some attempts to include spatial information in the network. 
Chen, Xinghao et al.\cite{chen2017pose} used local feature information to correct 3D coordinates. 
However, because it uses local segmentation to obtain local information, the network can obtain limited information and lack the information about the overall structure of the hand. 
Wu, X, etc.\cite{wu2018handmap} adopts a scheme combined with the detection-based method, 
which is similar to ours, but it needs to first convert the input depth map into a three-dimensional voxel map, 
and our method is to obtain the 3D coordinates directly in the plane space.

\subsection{Detection-based Method}
Comparing to the regression-based approach, detection-based methods\cite{ge2018robust, tompson2014real, chang2018v2v, wan2018dense} hopes use an intermediate SFR to maintain the spatial inforamtions, 
but often requires artificially design of SFR and lack direct supervision from the 3D coordinates. 
Since the method of directly applying heatmap can solve the problem of 2D coordinate recognition well, 
the detection-based method based on heatmap is very commonly used in human pose estimation. 
However, the detection-based method is not easy to implement in the hand pose estimation problem, 
because in addition to the need to identify the 2D coordinates in the image, it is also necessary to identify the depth coordinates, which is difficult to represent with a heatmap. 
However, due to the high precision and intuitiveness of the detection-based method, many scholars still try to use the Detection-based method in hand pose estimation. 
Tompson, J et al.\cite{tompson2014real} used a heatmap for 2D coordinate recognition, and then used a hand model-based iterative method PSO for post-processing to obtain depth coordinates. 
Gyeongsik Moon et al.\cite{chang2018v2v} converted the input depth image into a three-dimensional voxel map, and established a three-dimensional heatmap in the voxel map to convert the two-dimensional problem into a three-dimensional problem, and achieved good results. 
But this model relies on artificially defined pre-processing to complete the transformation from depth map to 3D voxel map. 
Wan, C et al.\cite{wan2018dense} proposed a method based on an offset vector field, which allows the neural network to learn the bias vector field, and then uses the mean-shift algorithm as decoder to convert the obtained vector field into 3D coordinates. 
Our design of SFRs is inspired by this work, but instead of bias vector field we use local offset map. We will compare with these models in the experiment part.

%
%

\section{METHODOLOGY}
\label{METHODOLOGY}

\begin{figure*}[!t]
  \centering
  \includegraphics[width=\linewidth]{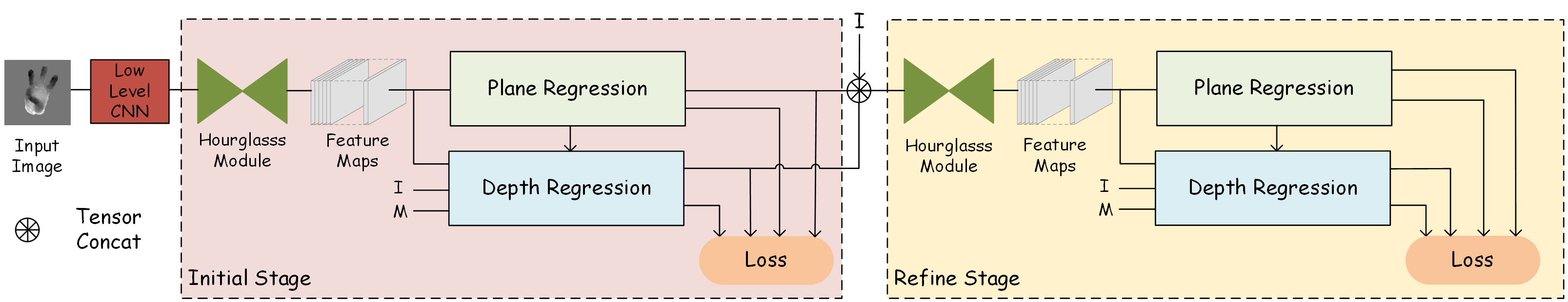}
  \caption{
  Overview of our designed model. 
  The input depth image goes through a low-level CNN to extract some low-level feature and down sampling to the representation scale. 
  The hourglass module is used to extract the overall features of the input image. 
  The Plane Regression module goes first after the feature extraction, 
  predicting the plane coordinates of each joint and also passing the predicted heat map to Depth Regression module. 
  On the other hand, the Depth Regression module, getting the heat map from Plane Regression module, 
  predicting the depth coordinates of each joint and also output the predicted depth map. 
  We concatenate the heatmap, depth map and representation scale image as the input of the refine stage whose structure is the same as the initial stage.
  }
  \label{model}
\end{figure*}

To make the problem formally and prevent confusion in the rest of the paper,   
the input of our model is a depth image $I_{D} \in \mathbb{R}^{n \times n}$, 
which only contains one single complete hand. 
That is our basic assumption. 
Some filter method and resize need to apply to the raw data to make such an image. 
We will discuss this in Section \ref{Experiment}. 
And the output of our model is the normalized UVD coordinates in the image plane of each joint $P_{j}^{j \in\{1 . . . J\}} \in \mathbb{R}^{3}$. 
Specifically, we separate the coordinates into two parts, $P_{j(u v)}$ and $P_{j(d)}$.
$P_{j(u v)} \in \mathbb{R}^{2}$ denote the plane coordinates in camera plane space. 
And $P_{j(d)} \in \mathbb{R}$ denote depth coordinates which is the distance between the camera and joint $j$. 
Notice that, we do not use XYZ coordinates as output like some model did, 
because the UVD coordinates are more direct information from a depth image without a transformation influenced by the internal parameters of the camera. 
In this section, we will first propose the Pixel-wise method, clarify the problem and why our method can help. 
Then we will design a particular model use Pixel-wise Regression. 
Specifically, we design the SFR and its correlative DD which consists of two module CR and DR. 
Finally, we show other techniques we use when building our model and some implement details about the network.

\subsection{Pixel-wise Regression method}

The 3D hand pose estimation problem from single depth image can be formalized as a function mapping from depth image to 3D joint coordinates, i.e. $\phi : I_{D} \rightarrow P$. 
The regression-based method just wants approximate this function with a single neural network. 
On the other hand, as we discuss above, the key motivation of detection-based method is that we can make a detour. 
That is instead of directly mapping to the joint coordinates we can first mapping to an encoded representation of it, i.e. $\varphi : I_{D} \rightarrow L$, 
then use a decoder to recover the joint coordinates, i.e. $f : L \rightarrow P$. 
In which $L$ denotes the encoded representation. 
To use the detection-based method, one must first define the encoder $g : P \rightarrow L$. 
Ideally, the encoder and the decoder are reciprocal, i.e. $P=f(g(P))$. 

The intuition here is quite simple, 
since the SFR and the direct supervision from 3D coordinates both benefit the result of the neural network model, 
we can just put the decoder inside the network and make use of both, just like what we shown in the Fig\@. \ref{three_types}. 
We call this type of method Pixel-wise Regression.

The essential requirement for this method is that the decoder $g$ must be differentiable. 
However, the decoder of the commonly used heatmap is $argmax$, 
which is a nondifferentiable operation and also not reciprocal of the complex encoder. 
The truth is, due to the complex nature of the encoder, which maps low-dimensional data to high dimension, 
the representation must contain certain redundancy which make it hard to perfectly recover the original data with a differentiable decoder. 
To use the Pixel-wise Regression method, we need to build a particular model. 
That is we need to design an SFR with a DD which is nontrivial. 
The overview of our model is shown in Fig\@. \ref{model}. 

\subsection{Design of SFR and DD}

As we discuss in Section \ref{Introduction}, since we are dealing with 3D coordinates in 2D plane, 
we need more than one form of SFR to encode the 3D coordinates. 
In our design, we use two SFRs, heat map and local offset depth map to encode plane coordinates and depth coordinates receptively. 
We assemble each SFR and its DD to module, thus we have two modules named Plane Regression and Depth Regression. 
The following of this part is to describe these two modules in details. 

\subsubsection{Plane Regression}

\begin{figure}[!t]
  \centering
  \includegraphics[width=2.5in]{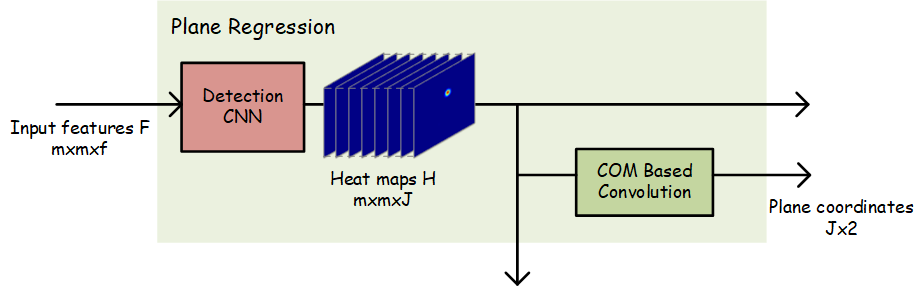}
  \caption{Detail structure of Plane Regression. 
  Input features go through a CNN network and predict the heatmap for each joint. 
  Then the heatmap goes through a COM(center of mass) based convolution to convert the heatmap to 2D coordinates.
  }
  \label{PR}
\end{figure}

We start from the easy part; the SFR of plane coordinates. 
The detailed structure of PR is shown in Fig\@. \ref{PR}. 
Generally, PR take a feature map $F \in \mathbb{R}^{m \times m \times f}$ as input and predict two things, 
a heatmap $\tilde{H}_{j} \in \mathbb{R}^{m \times m}$ and $\tilde{P}_{j(u v)}$ for each joint. 

What we need here is an SFR for plane coordinates with a DD. 
And because heatmap matches our human intuition and already achieve good result in 2D problem, 
we want to modify it to meet our requirement. 
The problem is that the normal heatmap use argmax as decoder which is nondifferentiable. 
A simple idea is to use the center of mass (COM) of the heatmap to represent the desire plane coordinates, i.e. 

\begin{equation}
  P_{j(u)}=\sum_{i, k} H_{j}(i, k) \cdot u(i, k)
\end{equation}

\begin{equation}
  P_{j(v)}=\sum_{i, k} H_{j}(i, k) \cdot v(i, k)
\end{equation}

where the $u(i,j)$ and $v(i,j)$ denote the normalized $u$, $v$ coordinates for the pixel $(i,j)$ .

This COM-based decoder has two clear advantages. 
First, it meets our requirement of being differentiable. 
Moreover, it is easy to make this decoder in a convolutional form, which is the most common building block in deep learning thus can accelerate computation. 
In the convolutional form, the above equation can be rewritten as: 

\begin{equation}
  P_{j(u v)}=H_{j} * C
\end{equation}

where $C \in \mathbb{R}^{m \times m \times 2}$ is the COM convolutional kernel, defined as: 

\begin{equation}
  C(i, j, k)=\left\{\begin{array}{ll}{u(i, j)} & {k=1} \\ {v(i, j)} & {k=2}\end{array}\right.
\end{equation}

To complete this module, we need to define an encoder to build our desired heatmap. 
Although the heatmap we want is unlike the commonly used heatmap, we still want to retain its unique intuitive nature, 
that is, the closer the pixel to the given joint is, the larger the value it will be. We divided the encoder into two steps. 
In the first step, we only take the four nearby pixels into account. 
Notice that it is still an indeterminate system because the COM in plane space can only provide two condition and we have four variables to solve. 
But since it is still a probability map, we can form some constraint as $H_{j}(i, k) \geq 0$ and $\sum_{i, k} H_{j}(i, k)=1$. 
And we choose the middle value of the efficient solution set as the four corners’ value for our basic heatmap. 
Then we apply a gauss kernel with size $k$ to the basic heatmap we get from the first step to form our desired heatmap. 
It is obvious that the convolution here does not change the COM for the given heatmap. 
The kernel size here measures how much redundancy we want to put into the heatmap. 
And in practice, the valid region, where the heatmap have a non-zero value, 
in the heatmap also measures the uncertainty that the network feels about the prediction of a certain joint, 
we will discuss more about this in Section \ref{Experiment}.

We use the mean-square-error between predicted value and ground truth as loss of this module 
which contains two part $L_{u v}$ and $L_{H}$ which denote the coordinates loss and representation loss respectively:

\begin{equation}
  L_{u v}=\sum_{j}\left\|P_{j(u v)}-\tilde{P}_{j(u v)}\right\|_{2}^{2}
\end{equation}

\begin{equation}
  L_{H}=\sum_{i, j, k}\left[H_{j}(i, k)-\tilde{H}_{j}(i, k)\right]^{2}
\end{equation}

\subsubsection{Depth Regression}

\begin{figure}[!t]
  \centering
  \includegraphics[width=2.5in]{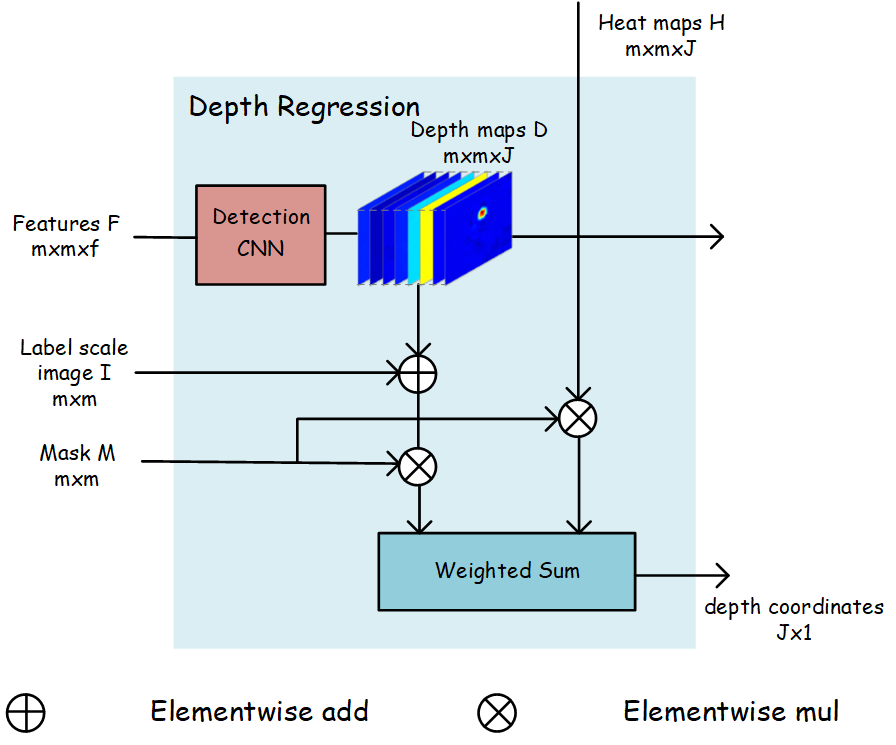}
  \caption{Detail structure of Depth Regression. 
  Input features go through a CNN network and predict the depth map for each joint. 
  Then combining with label scale image, we recover the depth information in local joint. 
  The mask filter values that do not on the hand. 
  Finally, a weighted sum is conducted to convert the heatmap and depth map to the depth coordinates.
  }
  \label{DR}
\end{figure}

Then we deal with the depth coordinates. 
The detailed structure of DR is shown in Fig\@. \ref{DR}. 
Generally, DR take a feature map $F \in \mathbb{R}^{m \times m \times f}$, 
a representation scale image $I \in \mathbb{R}^{m \times m}$ , 
a mask matrix $M \in \mathbb{R}^{m \times m}$ and the predicted heatmap $\tilde{H}_{j}^{j=\{1 \ldots J\}}$ from PR as input and predict two things, 
a local offset depth map $\tilde{D}_{j} \in \mathbb{R}^{m \times m}$ and $\tilde{P}_{j(d)}$ for each joint. 

As we discuss in the PR part, we need an SFR of depth coordinates with a DD. 
We have shown in Section I that the real depth coordinates of the joints have varied offsets for the values on the depth image. 
Therefore, the idea here is that we can use the neural network to predict these offsets in a spatial-form. 
Intuitively, like the heatmap, only the pixels that are close to the joints can give the information about this offset. 
So, we build a local offset depth map (depth map for short) to encode the depth information into plane space. 
The depth map is defined as:

\begin{equation}
  D_{j}(i, k)=\left\{\begin{array}{ll}{\left(P_{j(d)}-I(i, k)\right) \cdot M(i, k)} & {\text { if } H_{j}(i, k)>0} \\ {0} & {\text { otherwise }}\end{array}\right.
\end{equation}

We use the heatmap we build in the PR to define the local region of the given joint to make sure that the heatmap and the depth map have same level of redundancy. 
And the representation scale image $I$ is just a resized image from the input depth image $L_{D}$. 
The reason we need this image is that since we have some down sampling process in the network, the size of the depth map is different from the input depth image. 
Since we use the depth offset to encode the depth coordinates, it is meaningless if there are some values not on the hand. 
The offset value for the background is just the exact value for the depth coordinates of the joints, which makes the problem no easier than the direct regression. 
Hence, we use mask matrix $M$ which denotes which pixel is on the hand to filter the depth map. 
The mask matrix $M$ is defined as:

\begin{equation}
  M(i, k)=\left\{\begin{array}{ll}{1} & {\text { if } I(i, k)>0} \\ {0} & {\text { otherwise }}\end{array}\right.
\end{equation}

After we design the SFR of depth coordinates, we also need a DD to fuse the redundant depth map to a scalar. 
Ideally, we can fully recover the depth coordinates by adding any two value of $D$ and $I$ in the local region of the given joint. 
So, it seems that the DD can be simply defined as averaging the recover values. 
However, since the depth map $\tilde{D}_{j}$ is predicted by the neural network, the result will not be perfect. 
We want the depth map has some same property as heatmaps, which has more accuracy when closer to the given joint. 
Therefore, our decoder uses the heatmap to weight the recover values given by the neural network:

\begin{equation}
  \tilde{P}_{j(d)}=\frac{\sum_{i, k} M(i, k) \tilde{H}_{j}(i, k)\left[I(i, k)+\tilde{D}_{j}(i, k)\right]}{\sum_{i, k} M(i, k) \tilde{H}_{j}(i, k)}
\end{equation}

Due to the background problem we state above, we also use the mask matrix M here to ignore the pixel off the hand.
 
Like the PR, we use the mean-square-error as well. 
The loss of this module also contains two part $L_{d}$ and $L_{D}$ which denote the coordinates loss and representation loss respectively:

\begin{equation}
  L_{d}=\sum_{j}\left(P_{j(d)}-\tilde{P}_{j(d)}\right)^{2}
\end{equation}

\begin{equation}
  L_{D}=\sum_{i, j, k}\left[D_{j}(i, k)-\tilde{D}_{j}(i, k)\right]^{2}
\end{equation}

\subsection{Network architecture}

As shown in previous works\cite{wei2016convolutional, chen2017pose, wan2018dense, bulat2016human}, 
the multi-stage network can refine the result and 
deal with occlusions by inferring from results from the previous stage. 
In our model, making a tradeoff between speed and accuracy, we use a typically two-stage network, which uses the same structure. 
The loss function of the multi-stage network can be defined as the sum loss of each stage:

\begin{equation}
  L=\sum_{s} L^{(s)}
\end{equation}

In our model, the loss function for a single stage is the sum of loss of PR and DR:

\begin{equation}
  L^{(s)}=L_{i w}^{(s)}+L_{d}^{(s)}+\lambda_{H} L_{H}^{(s)}+\lambda_{D} L_{D}^{(s)}
\end{equation}

The $\lambda_{H}$ and $\lambda_{D}$ are two scale factors to make sure the losses are on a similar scale. 
In our model, we empirically set $\lambda_{H} = \lambda_{D} = 1$.

We also use the hourglass module\cite{newell2016stacked} as our main feature extractor. 
The hourglass module, due to its unique recursive structure, can provide features extracted in multi-scales and 
widely used in pose estimation problems. 

We have two sizes of the image in our model. 
Only the input depth image is in a bigger size of $\mathbb{R}^{m \times m}$, 
and all the other inputs and representations are in the size of $\mathbb{R}^{n \times n}$. 
We use a bigger image to contain more details in the low-level features as we have a low-level CNN block before the initial stage 
which extracts low-level features and down sampling it to $\mathbb{R}^{n \times n}$.

\subsection{Implement Detial}

We use TensorFlow framework to build and train our network using Adam optimizer with a learning rate of $2e-4$. 
Since our model uses massive channel-wise operation, we use instance normalization\cite{ulyanov2016instance} to accelerate training. 
We use a similar data normalization method to\cite{oberweger2017deepprior}, 
which extracts a fixed-size cube centered on the center of mass of this object from the depth image. 
We set the input image size $m$ to 128, and representation size $n$ to 64. 
We also implement a [-30, 30] random rotation as data augmentation. 
Since our model is an FCN which is translation invariant, we do not perform any random translation for data augmentation. 
Empirically, we set the kernel size $k$ to 7 to get certain level of redundancy. 
The batch size is set to 32, and we train our model with 10 epochs. 
In the test time, our model can achieve about 100 FPS on a single TITAN XP GPU.

\section{Experiment}
\label{Experiment}

In this section, we conduct experiments on three challenge public datasets, i.e. MSRA, ICVL and HAND17. 
We choose the MSRA dataset to conduct the ablation experiment because the dataset contains only full hand image 
that fit our assumption for the input image best. 
We used two metrics in all experiments. The first is the mean 3D distance error for all joint points. 
Another metric, which is more challenging, is that the percentage of frames in which all joint errors are below a certain threshold.

\subsection{Datasets}

\subsubsection{MSRA dataset}

The MSRA hand pose dataset\cite{sun2015cascaded} contains 76500 frames from 9 different subjects captured by Intel’s Creative Interactive Camera. 
The dataset provides annotations for 21 joint points, four joints per finger, and one joint on the palm. 
Since the dataset itself does not divide the training set and the test set, 
we randomly divide the training set and the test set by a ratio of 8:1. 
Since the data itself is a segmented hand image, we do not conduct any new segmentation.

\subsubsection{ICVL dataset}

The ICVL dataset\cite{tang2014latent} contains 330k frames from 10 different subjects also using Intel’s Creative Interactive Gesture Camera. 
This dataset contains annotations for 16 joint points, three for each finger and one for the palm. 
The dataset contains random rotations, and since we used online data augmentation, we only used 22k of original images. 
Since the image contains the background, we first use a boundary box built by the ground truth of joints coordinates to 
roughly remove the background. Then we use an empirical threshold to remove the rest.

\subsubsection{HAND17 dataset}

The HAND17 dataset\cite{yuan2018depth} is the largest scale dataset in 3D hand pose estimation field 
which contains 957k frames for training and 295k frames for testing. 
The dataset captures the depth images by the latest Intel RealSense SR300 camera and 
automatic annotates 21 joints using six 6D magnetic sensors and inverse kinematics. 
Since the images provided by the dataset has background, we use the same procedure we describe in ICVL dataset to remove them. 
The test set provides the boundary box for us, thus we only use the empirical threshold.

\subsection{Ablation Experiment}

\begin{figure}[!t]
  \centering
  \includegraphics[width=2.5in]{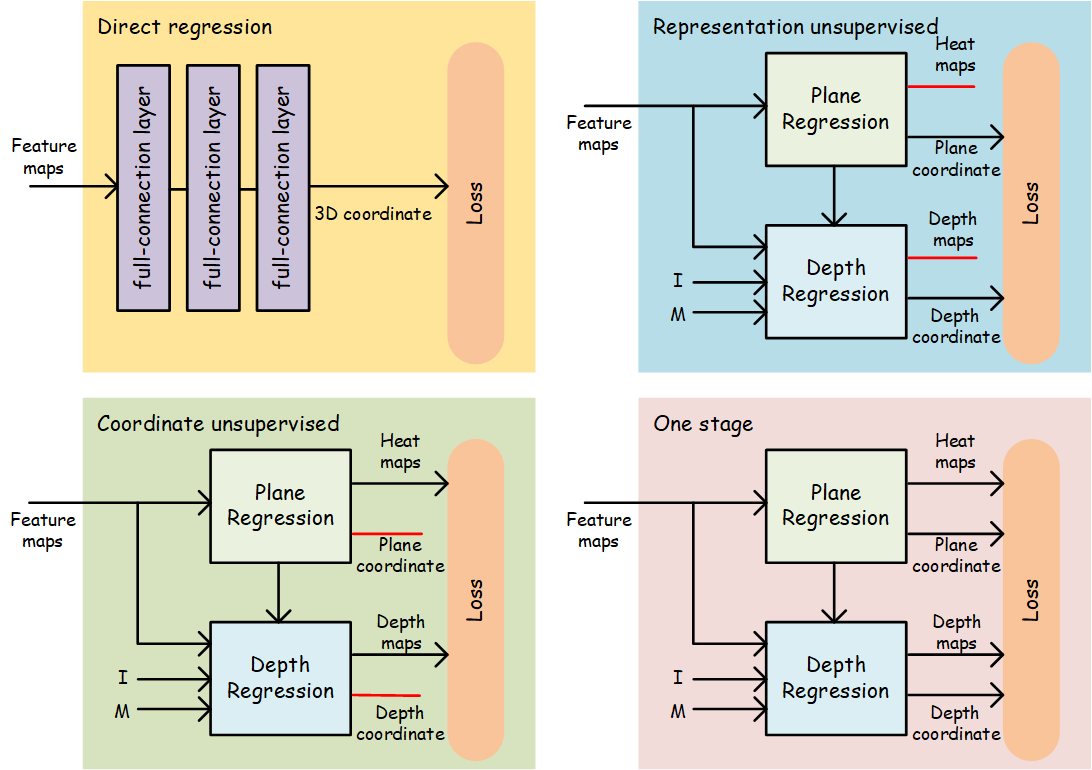}
  \caption{Setup of the ablation experiment.}
  \label{Ablation_setup}
\end{figure}

To demonstrate the validity of our method, we design and perform an ablation experiment. 
As shown in Fig\@. \ref{Ablation_setup} we modify some structures in our model to form four different baseline models. 
To set up a fare comparison, we retain the feature extraction part of the model and take the extracted features as the input of the ablation experiment. 
Generally, we conduct this ablation experiment for three purposes. 

First, we want to prove that the Pixel-wise Regression method we proposed is better than the former two methods. 
In order to compare with the regression-based method, 
we replace PR and DR with three fully connected layers and directly regress 3D coordinates from features. 
We call this model Direct\_Rregression. 
On the other hand, in order to compare with the detection-based method, 
we remove the supervision of the joint coordinates and put the decoder outside the neural network. 
We call this model Coordinate\_Unsupervised. 

Second, we want to explore how different training strategies influence the learned SFRs and results. 
As we discuss in Section III, although we design reasonable SFRs, 
we control the redundancy in such representation by an empirical kernel size $k$. 
In the original design we use, we supervised the coordinates along with the representation, 
which can guide the representation predicted by the network converge to the representation we design. 
However, even if we do not supervise the representation, 
the model can still learn a representation of 3D coordinates due to the gradient backpropagated by the decoder. 
The decoder here serves not only a regression module but also a constraint that limits the search space of the representation. 
Because there are infinite such SFRs, the learning outcomes of the network may be different from supervised learning model. 
To check out the outcome of such a model and get deeper understanding of our model, 
we remove the supervision for the representation and call this network Representation\_Unsupervised. 

Finally, we want to verify that the multi-stage network is able to improve the accuracy of the predictions. 
So, in the last comparison experiment we only keep the first stage, so we call this model One\_Stage. 
We compare these four models with our designed two-stage model which we call Two\_Stage here on the MSRA dataset.

\begin{figure}[!t]
  \centering
  \includegraphics[width=2.5in]{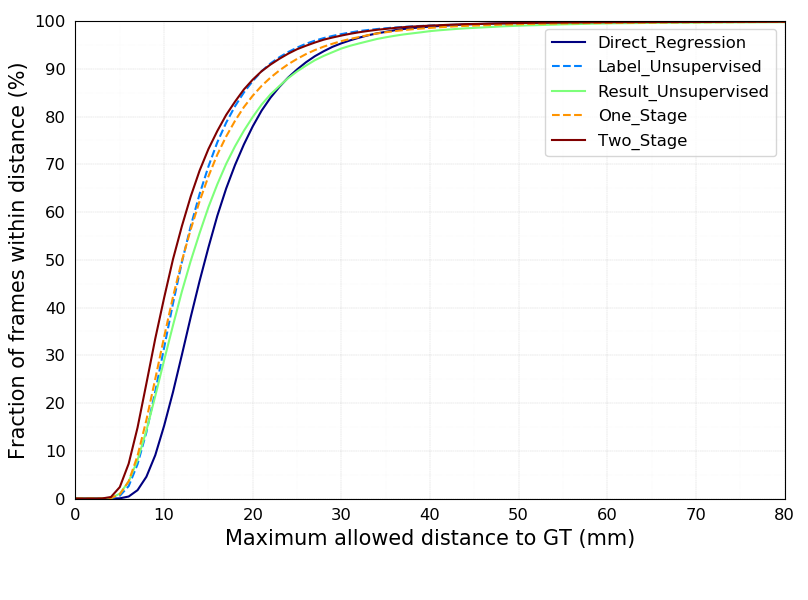}
  \caption{The result of the ablation experiment using the maximum allowed distance metric.}
  \label{Ablation_Plot}
\end{figure}

The results of the ablation experiment are shown in Fig\@. \ref{Ablation_Plot}. 
We can see that as we estimated, Direct\_Regression, which does not preserve spatial information, yields the worst results. 
Coordinate\_Unsupervised, a detection-based method, yields a better result than the regression-based method. 
However, since there is no direct supervision from coordinates, the two modules are learned separately and cannot perfectly cooperate, so the result is not good either. 
Moreover, compared with the One\_Stage model, it can be found that the Two\_Stage method does have a significant improvement in the accuracy. 

\begin{figure}[!t]
  \centering
  \includegraphics[width=2.5in]{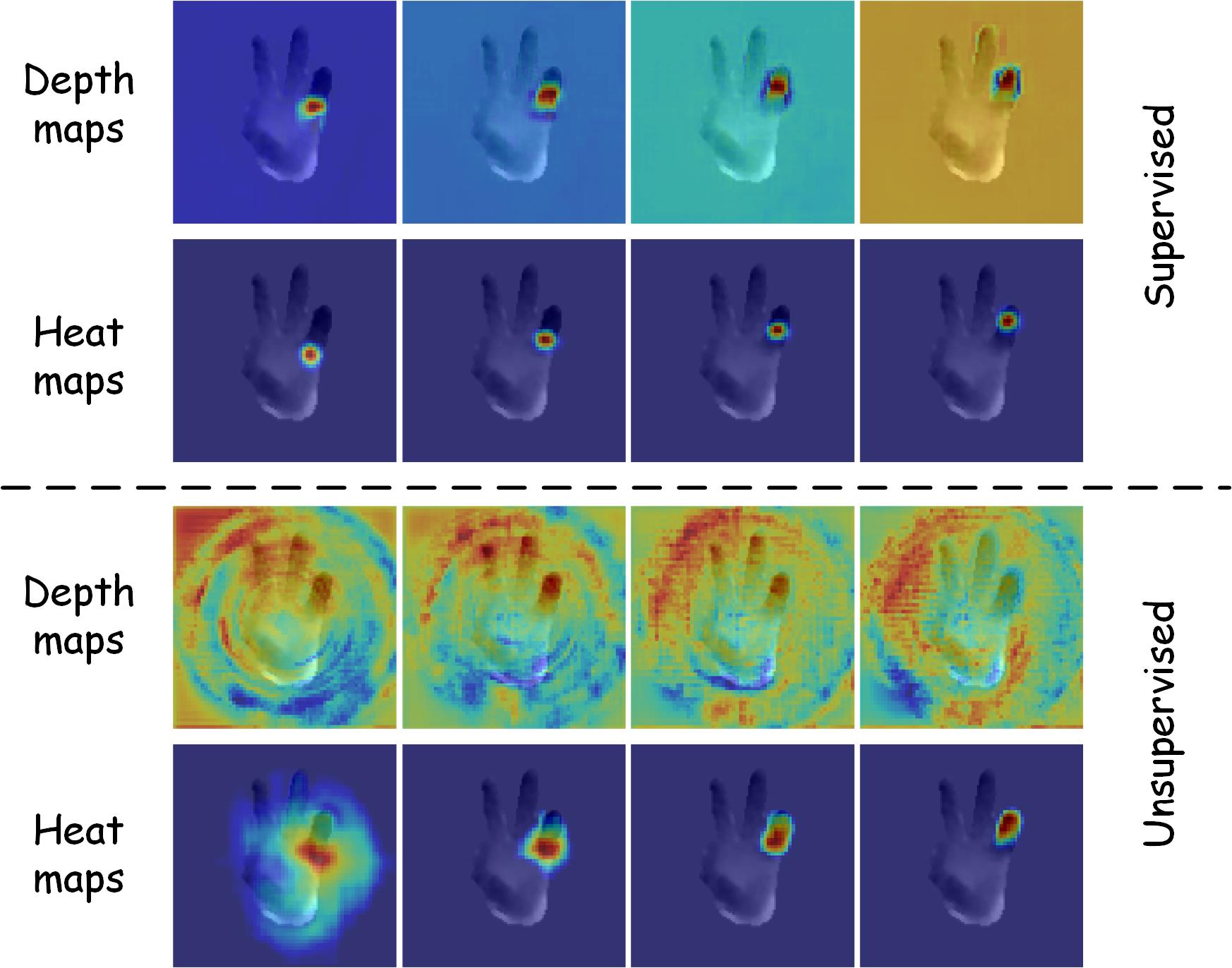}
  \caption{Learned equivalent representation. 
  The top is the result of supervised model, and the bottom is the result of unsupervised method. 
  For each part, the first row is learned depth map, and the second row is learned heatmap. 
  From left to right is MCP, PIP, DIP, TIP joint of the index finger receptively. Best view in color.
  }
  \label{Ablation_Image}
\end{figure}

For the Representation\_Unsupervised method, we can find that its performance is better than other methods, 
and yield comparable result with our original design. 
Since we want to study the learned representation effect by supervision, 
we draw the learned representation upon the input depth image, as shown in Fig\@. \ref{Ablation_Image}.  
The result is quite interesting. 
We can see that the supervised representation tends to have a low-level redundancy as we design and have same level of redundancy for different joints. 
The valid region of depth map is slightly larger than the heatmap, but they converge to the same joint. 
That is because the recovered depth values given by depth map are weighted by the heatmap which cannot be precisely predicted. 
Therefore, the depth map must make the prediction in a larger local region to compensate the uncertainty of the heatmap. 
This is also what make our Pixel-wise Regression method better than the detection-based method. 

On the other hand, the unsupervised model has more redundancy for the larger joints, i.e. the MCP joint. 
That is because the larger the joint the more similar the local region looks like. 
For instance, the MCP joint is closer to the palm and other root joint of finger, so it is always surround by the dense and similar information provided by the depth image. 
Thus, without the supervision, these similarities can cause confusion and result in uncertainty. 
On the other hand, the TIP joint, which is usually locates at the boundary of the hand has a clearer context and has lower uncertainty. 
The depth map predicted by the unsupervised model is quite strange with a circular form and do not focus on a particular region. 
That is because the valid region of depth map is defined by heatmap and mask matrix, when we do not supervise the depth map, 
there are zero gradients for those invalid pixels which gives the network more freedoms to choose the value for them. 
These freedoms loosen the local constrain when we design the depth map and may cause the model to overfit to particular patterns. 
Actually, the circular form of the depth map is an evidence that the network is overfit to the random rotation we perform during data processing.

\subsection{Comparing with State-of-the-Art}

In this part, we compare our model with the state-of-the-art models. 
We have chosen models that are representative and have the state-of-the-art accuracy. 
In addition to the models we mentioned in Section II, we also include 
Region Ensemble network (include REN-9x6x6\cite{wang2018region} and REN-4x6x6\cite{guo2017region}), 
3D Convolutional Neural Networks (3DCNN\cite{ge20173d}), 
SHPR-Net\cite{chen2018shpr}, 
HandPointNet\cite{ge2018hand}, 
Point-to-Point\cite{ge2018point}. 
We use tools provide by Chen, X\cite{chen2017pose} to evaluate the result.   

\begin{figure*}[!t]
  \centering
  \includegraphics[width=\linewidth]{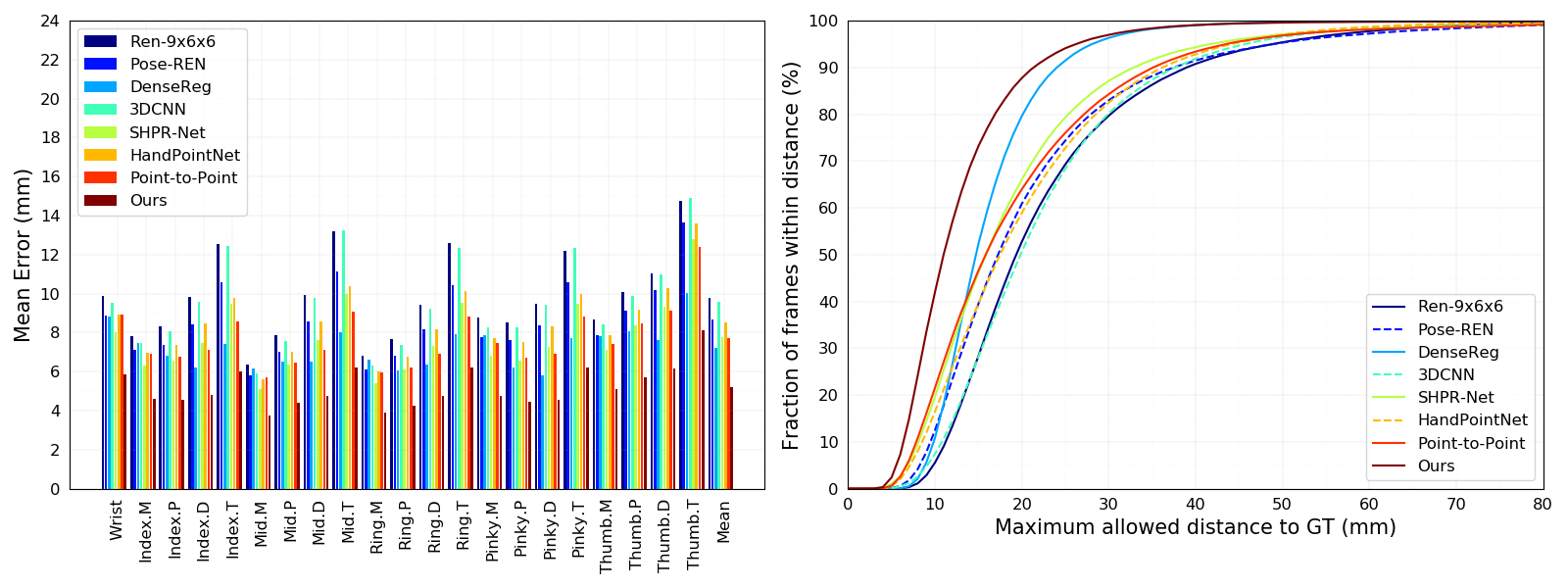}
  \caption{
    The results of MSRA dataset. 
    Left is mean error (mm) for each joint. 
    Right is the proportion of frame that all joints error are under the given threshold.
  }
  \label{msra_error}
\end{figure*}

\begin{table}[!t]
  \renewcommand{\arraystretch}{1.3}
  \caption{MEAN 3D ERROR ON MSRA DATASET}
  \label{msra_table}
  \centering
    \begin{tabular}{c c}
    \toprule
    Model & 3D error (mm)\\
    \hline
    REN-9x6x6\cite{wang2018region} & 9.792\\
    Pose-REN\cite{chen2017pose} & 8.649\\
    DenseReg\cite{wan2018dense} & 7.234\\
    3DCNN\cite{ge20173d} & 9.584\\
    SHPR-Net\cite{chen2018shpr} & 7.756\\
    HandPointNet\cite{ge2018hand} & 8.505\\
    Point-to-Point\cite{ge2018point} & 7.707\\
    \hline
    Ours & 5.186\\
    \bottomrule
    \end{tabular}
\end{table}

As shown in Fig\@. \ref{msra_error} and TABLE \ref{msra_table}. 
in MSRA dataset, our model greatly outperforms current state-of-the-art models in all joints. 
For the mean error of all joints, we only have 75\% of the best results present\cite{wan2018dense}, which is a big improvement. 
In addition, as can be seen in Fig\@. \ref{msra_error}.  our model greatly exceeds other models with a small allowable threshold, 
which means that we propose that the model is very effective in high-precision recognition.

\begin{figure*}[!t]
  \centering
  \includegraphics[width=\linewidth]{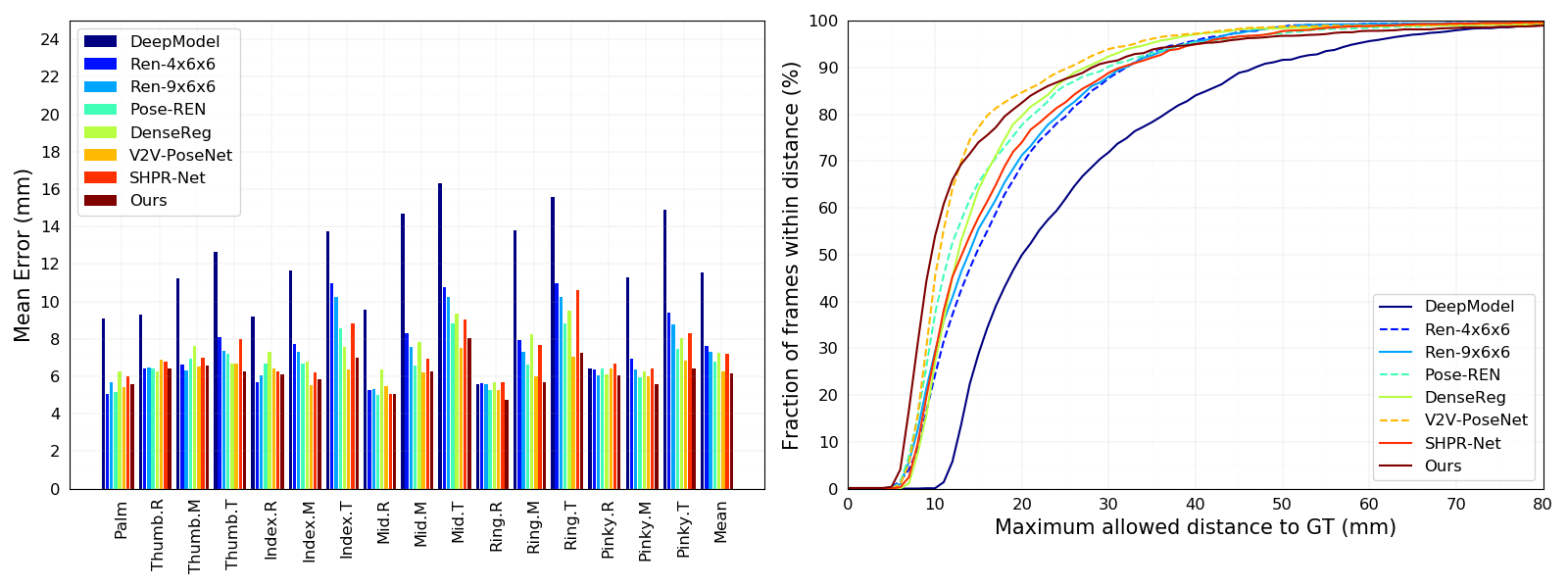}
  \caption{
    The results of ICVL dataset. 
    Left is mean error (mm) for each joint. 
    Right is the proportion of frame that all joints error are under the given threshold.
  }
  \label{icvl_error}
\end{figure*}

\begin{table}[!t]
  \renewcommand{\arraystretch}{1.3}
  \caption{MEAN 3D ERROR ON ICVL DATASET}
  \label{icvl_table}
  \centering
    \begin{tabular}{c c}
    \toprule
    Model & 3D error (mm)\\
    \hline
    DeepModel\cite{zhou2016model} & 11.561\\
    REN-4x6x6\cite{guo2017region} & 7.628\\
    REN-9x6x6\cite{wang2018region} & 7.305\\
    Pose-REN\cite{chen2017pose} & 6.791\\
    DenseReg\cite{wan2018dense} & 7.239\\
    SHPR-Net\cite{chen2018shpr} & 7.219\\
    V2V-PoseNet\cite{chang2018v2v} & 6.284\\
    \hline
    Ours & 6.177\\
    \bottomrule
    \end{tabular}
\end{table}

The results of ICVL dataset are shown in Fig\@. \ref{icvl_error} and TABLE \ref{icvl_table}.  
As can be seen from the results, we have achieved similar results with the best model\cite{chang2018v2v} on the average 3D error. 
As can be seen from Figure \ref{icvl_error}, when the error threshold is less than 10mm, we exceed all other models, 
but when the threshold is greater than 30mm, our model is not as effective as other models. 
This is because we manually segmented the image and the background of the image was not completely removed, 
which affects the performance of our model.

\begin{table}[!t]
  \renewcommand{\arraystretch}{1.3}
  \caption{MEAN 3D ERROR ON HAND17 DATASET}
  \label{Hand17_table}
  \centering
    \begin{tabular}{c c c c}
    \toprule
    Team Name & Average (mm) & Seen (mm) & Unseen (mm)\\
    \hline
    BUPT & 8.39	& 6.06 & 10.33\\
    SNU CVLAB & 9.95 & 6.97 & 12.43\\
    NTU & 11.30 & 8.86 & 13.33\\
    THU VCLab & 11.70 & 9.15 & 13.83\\
    NAIST\_RV & 12.74 & 9.73 & 15.24\\
    HuPBA & 14.74 & 11.87 & 17.14\\
    NAIST RVLab G2 & 16.61 & 13.53 & 19.18\\
    Baseline & 19.71 & 14.58 & 23.98\\
    \hline
    Ours & 12.22 & 8.73 & 15.13\\
    \bottomrule
    \end{tabular}
\end{table}

The result of HAND17 dataset is shown in TABLE \ref{Hand17_table}. 
Although it is two to three millimeters worse than the best model, we still achieve comparable result with other competitors. 
Specially, we find that our model gets better performance on seen joints than the unseen ones. 
We attribute this to the cause that our design of SFR is not efficient for the self-occlusion joints which have larger variation range than the seen ones. 

\begin{figure*}[!t]
  \centering
  \includegraphics[width=\linewidth]{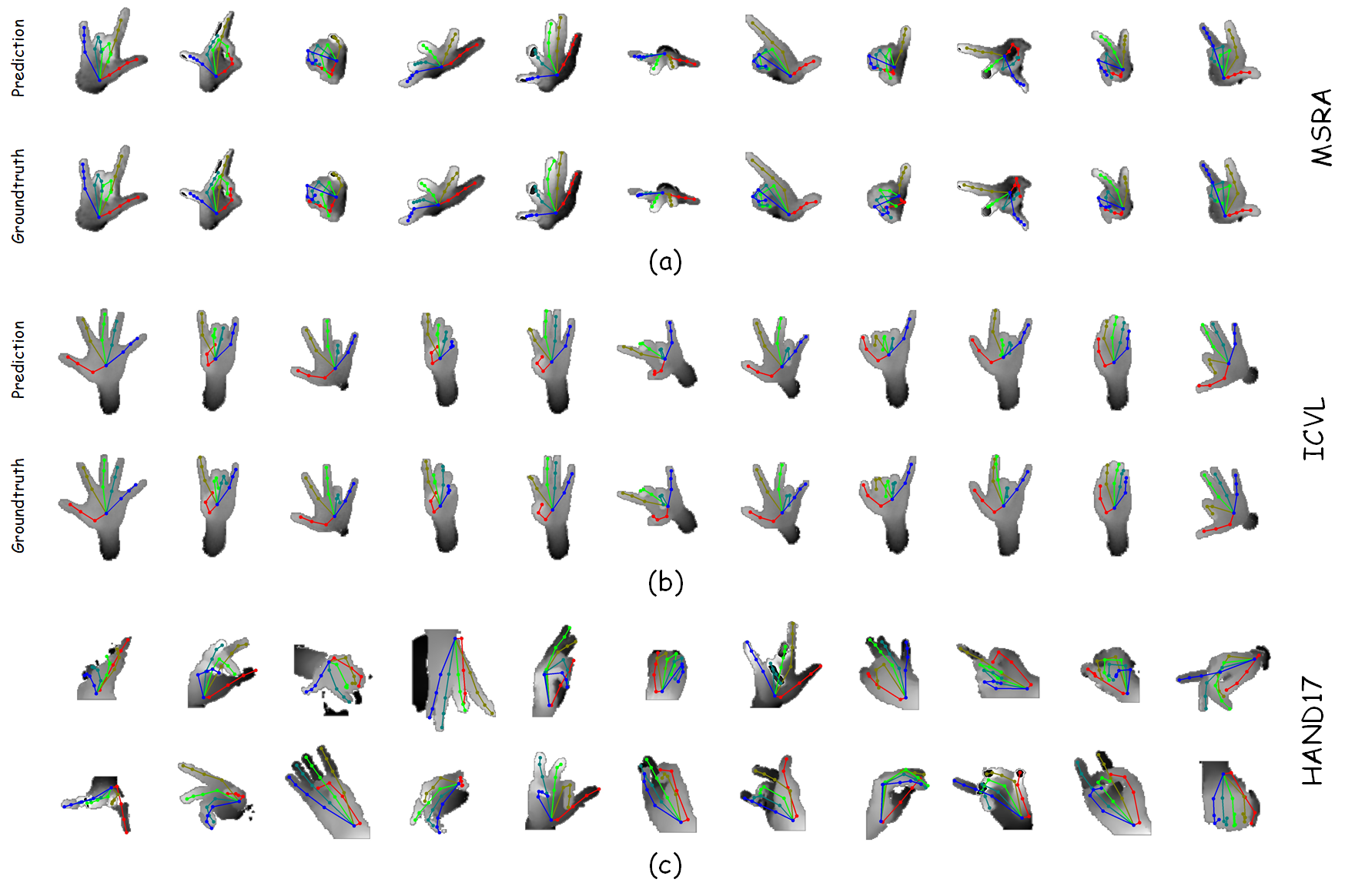}
  \caption{
    Qualitative results of our model. (a), (b), (c) shows the result form MSRA, ICVL, HAND17 dataset respectively. 
    Specially, in (a) and (b) we provide ground truth in the second row for comparison.
  }
  \label{qualitative_result}
\end{figure*}

We also show some qualitative results in these two datasets in Fig\@. \ref{qualitative_result}. 
Some of these results are even better than the original annotations, 
which demonstrate that our model learn the essential information of the data and not just remember all the patterns.

\section{Conclusion}
\label{Conclusion}

In this paper, we propose a Pixel-wise Regression method for 3D hand pose estimation, 
which use spatial-form representation and differentiable decoder to solve the losing spatial information and 
lacking direct supervision problems faced by existing methods. 
We design a particular model that use our proposed method. 
Specifically, we design the spatial-form representation and its correlative differentiable decoder 
which consists of two modules Plane regression and Depth Regression 
that deal with plane coordinates and depth coordinates respectively. 
The ablation experiment shows that our proposed method is better than the former methods. 
And it also shows that the supervision on the representation is vital to the performance for our model. 
Experiments on public datasets show that our model reach the state-of-the-art level performance. 

In the future, we intend to explore more design of representation used for Pixel-wise Regression method or 
directly use our model in the field of human-computer interaction to control real robotics hand.


%





\ifCLASSOPTIONcaptionsoff
  \newpage
\fi



\bibliographystyle{IEEEtran}
\bibliography{IEEEabrv, reference}
\end{document}